# The Impact of Automatic Pre-annotation in Clinical Note Data Element Extraction — the CLEAN Tool

Tsung-Ting Kuo, PhD[1], Jina Huh, PhD[1], Jihoon Kim, MS[1], Robert El-Kareh, MD, MS, MPH[1], Siddharth Singh, MD[1], Stephanie Feudjio Feupe, MSc[1], Vincent Kuri, MS[2], Gordon Lin, MS[2], Michele E. Day, PhD[1], Lucila Ohno-Machado, MD, PhD[1], and Chun-Nan Hsu, PhD[1, *]

[1] UCSD Health Department of Biomedical Informatics, University of California San Diego, La Jolla, CA, USA

[2] Department of Computer Science and Engineering, University of California San Diego, La Jolla, CA, USA

[*] 9500 Gilman Dr, San Diego, CA 92093, USA; chunnan@ucsd.edu; +1 (858) 822-4931.

**ABSTRACT**

**Objective**. Annotation is expensive but essential for clinical note review and clinical natural language processing (cNLP). However, the extent to which computer-generated pre-annotation is beneficial to human annotation is still an open question. Our study introduces CLEAN (CLinical note rEview and ANnotation), a pre-annotation-based cNLP annotation system to improve clinical note annotation of data elements, and comprehensively compares CLEAN with the widely-used annotation system Brat Rapid Annotation Tool (BRAT).

**Materials and Methods**. CLEAN includes an ensemble pipeline (CLEAN-EP) with a newly developed annotation tool (CLEAN-AT). A domain expert and a novice user/annotator participated in a comparative usability test by tagging 87 data elements related to Congestive Heart Failure (CHF) and Kawasaki Disease (KD) cohorts in 84 public notes.

**Results**. CLEAN achieved higher note-level F1-score (0.896) over BRAT (0.820), with significant difference in correctness (P-value < 0.001), and the mostly related factor being system/software (P-value < 0.001). No significant difference (P-value 0.188) in annotation time was observed between CLEAN (7.262 minutes/note) and BRAT (8.286 minutes/note). The difference was mostly associated with note length (P-value < 0.001) and system/software (P-value 0.013). The expert reported CLEAN to be useful/satisfactory, while the novice reported slight improvements.

**Discussion**. CLEAN improves the correctness of annotation and increases usefulness/satisfaction with the same level of efficiency. Limitations include untested impact of pre-annotation correctness rate, small sample size, small user size, and restrictedly validated gold standard.

**Conclusion**. CLEAN with pre-annotation can be beneficial for an expert to deal with complex annotation tasks involving numerous and diverse target data elements.

1. **BACKGROUND AND SIGNIFICANCE**

Clinical notes with unstructured narrative, such as progress notes, radiology reports, and discharge summaries, are one of the most information-rich, under-utilized sources of healthcare data.[1] Critical aspects of clinical quality are often described in the free-text notes of electronic health records (EHR) systems. These important aspects can be used to improve healthcare delivery/management, clinical/translational research, and ultimately patient health.

*Clinical Natural Language Processing (cNLP)* is dedicated to developing tools and systems to extract such useful information from medical text. Widely used cNLP tools include *cTAKES* (clinical Text Analysis and Knowledge Extraction System),[2] *MetaMap*,[3] *MedEx*,[4] *CLAMP* (Clinical Language Annotation, Modeling, and Processing Toolkit),[5] *Vitals*,[6] *EFEx*,[7] *KD-NLP*,[8] and a few others.[9-11] These cNLP tools can extract various types of information such as *condition* and *medication*.

*Annotation* in cNLP refers to the process of manually identifying the mentions of data elements of target signs, symptoms, events, *etc.* to be extracted in clinical notes. Annotation, although imperfect, is an important process that provides: (1) quality control for final cNLP output data, (2) gold standards to evaluate the performance of cNLP tools, and (3) training examples to develop and improve cNLP tools. Specifically, training examples are essential for those cNLP tools based on supervised machine learning, as well as for the development of rule-based cNLP tools.[9-20]

Annotation, however, is also the bottleneck of the whole development process of cNLP tools,[21] especially when numerous and diverse data elements are to be extracted.[1] From our experience in the patient-centered SCAlable National Network for Effectiveness Research (pSCANNER) project,[22] an experienced clinical annotator required an average of 15 to 30 minutes to annotate a clinical note for an annotation task involving the tagging of 41 data elements.

Intuitively, *pre-annotation* by a cNLP tool before a manual review might help improve the correctness and efficiency of the annotation process.[12] In pre-annotation, the mentions of the target data elements are identified by a cNLP tool. These elements serve as suggestions to a human annotator, so that the annotator can review and revise the pre-annotated mentions ("*pre-annotations*") instead of starting the annotation process from scratch. However, previous cNLP studies of pre-annotation showed mixed and inconsistent results in terms of correctness and efficiency on tasks that included name entity recognition,[12 13] de-identification,[14 15] patient records chart review,[16 17] corpus creation,[18] and NLP output validation,[19 20]. None of the studies considered the tailored design of a user interface (UI) to take advantage of pre-annotation. In the study of name entity recognition, some authors reported positive results of pre-annotation [12]: *"Time savings [of pre-annotations] ranged from 13.85% to 21.5% per entity. Inter-annotator agreement (IAA) ranged from 93.4% to 95.5%. … The time savings were statistically significant. Moreover, the pre-annotation did not reduce the IAA or annotator performance."* However, other authors provided contrasting outcomes [13]: *"We found little*

*benefit to [pre-annotate] the corpus with a third-party name entity recognizer ... the annotators (who were] given the [MetaMap Transfer (MMTx) tool, now MetaMap] [23] annotations, A1 and A3, annotated slower than the other two annotators, A2 and A4. ... There was also no clear trend that the MMTx annotation improved pair-wise IAA between individuals."* Most studies only focused on a limited number of target data elements to be extracted. For example, the authors of these studies [12] [13] named 5 and 10 concepts, respectively.

## 2. OBJECTIVE

Our goal was to take advantage of pre-annotation to improve the annotation process of clinical notes when numerous and diverse data elements that are necessary to support phenotyping and cohort identification were involved. In other words, we sought to design, implement and evaluate a software system that leverages pre-annotation to improve the annotation quality of clinical notes with a large number (> 50) of target data elements. This situation commonly occurs in cNLP use cases for phenotyping and cohort identification.[24-26] To this end, we developed the <u>CL</u>inical note r<u>E</u>view and <u>AN</u>notation (CLEAN) cNLP system and comprehensively compared CLEAN with the widely-used annotation system Brat Rapid Annotation Tool (BRAT).[21]

## 3. MATERIALS AND METHODS

### 3.1 The CLEAN cNLP System

The overall architecture of the CLEAN cNLP system is shown in Figure 1. The input to CLEAN includes a study plan specifying the target disease/condition, categorized data elements, and criteria used to select a set of clinical notes from an institute electronic health record (EHR) data warehouses. Since clinical notes contain PHI, we deposit them in a secure, privacy-preserving computation environment that also stores data elements definitions and categories.

The *CLEAN ensemble pipeline (CLEAN-EP)* pre-annotates the clinical notes using *Union*,[1] which extracts the data element if any of the cNLP tools identifies the data element and uses the ensemble method to integrate four cNLP tools (*cTAKES*,[2] *MetaMap* [3],*MedEx*,[4] and *KD-NLP* [8]), as well as our newly developed cNLP tool *ROCK (Rules for Obesity, Congestive heart failure, and Kawasaki disease)* described in Appendix A.1. We previously discovered that *Union* showed the highest recall/sensitivity among all of the ensemble methods in our pilot study[1]. Thus, when using *Union* as the ensemble method, annotators are provided with a high-coverage pre-annotated data element mentions in a clinical note text, instead of searching through the full text to identify any missing elements. That is, annotators can focus on correcting the pre-annotations, instead of adding the missing mentions. A benefit of applying an ensemble pipeline like CLEAN-EP is improved system flexibility to reuse and share cNLP tools.[27-32] Details about CLEAN-EP are described in Appendix A.1.

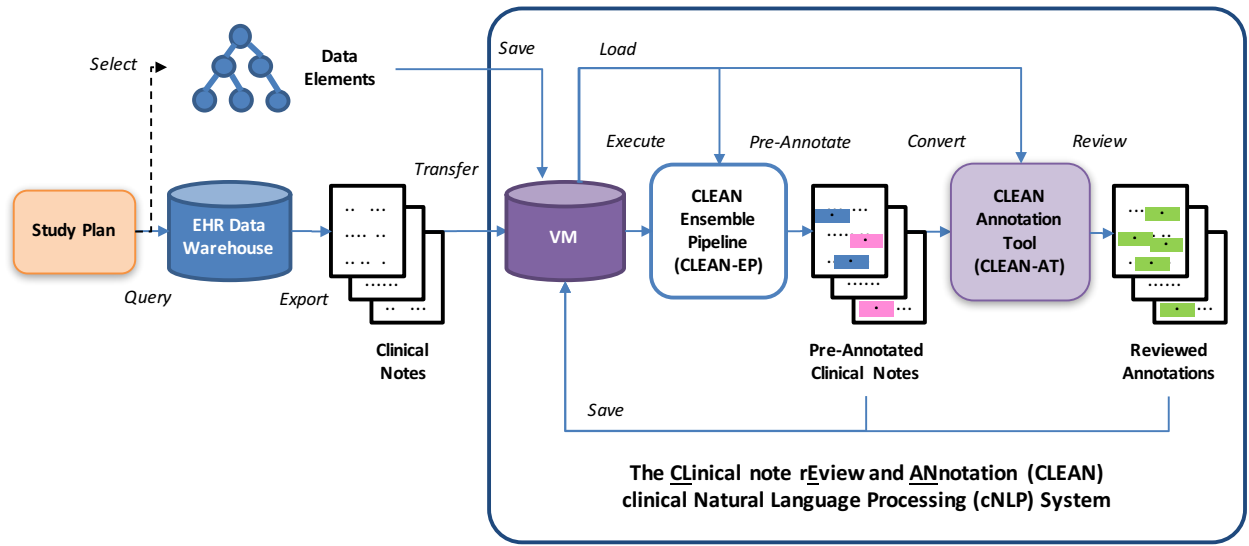

**Figure 1.** The <u>CL</u>inical note r<u>E</u>view and <u>AN</u>notation *(CLEAN)* clinical Natural Language Processing (cNLP) System. After the study plan was confirmed, the inputs were the selected data elements and the queried clinical notes from the electronic health record (EHR) data warehouse, both stored on a Virtual Machine (VM). Then, the built-in *CLEAN ensemble pipeline (CLEAN-EP)* pre-annotated the notes, followed by the user's review using the *CLEAN annotation tool (CLEAN-AT)*. The outputs were the reviewed annotations of the data elements on the clinical notes, and CLEAN stored the final results back on the VM.

Finally, an annotator will review and correct the pre-annotations using the *CLEAN annotation tool (CLEAN-AT)*. CLEAN-AT saves the final resulting annotations and allows annotators to re-check and refine the annotations if required. Details of CLEAN-AT are given in Section 3.2.

**3.2 The CLEAN Annotation Tool (CLEAN-AT)**

The UI design of the CLEAN-AT is illustrated in Figure 2, with the following main features:

- *Annotation editor (right panel)*. This main working space of CLEAN-AT shows the clinical note as well as all annotations on the note. A user can edit the annotations and review results on this panel. In this study, the terms *user* and *annotator* are used interchangeably.

- *Data element browser (left panel)*. Clicking the data element or category in the left panel quickly identifies all annotated mentions of that data element or category of the clinical note shown in the annotation editor. In this example (Figure 2), "Natriuretic Peptides" was clicked in the data element browser to identify the synonym mention "BNP" in the clinical note text. Conversely, in the clinical note, the mentions could be clicked in the annotation editor to highlight the corresponding data element in bold in the data element browser.

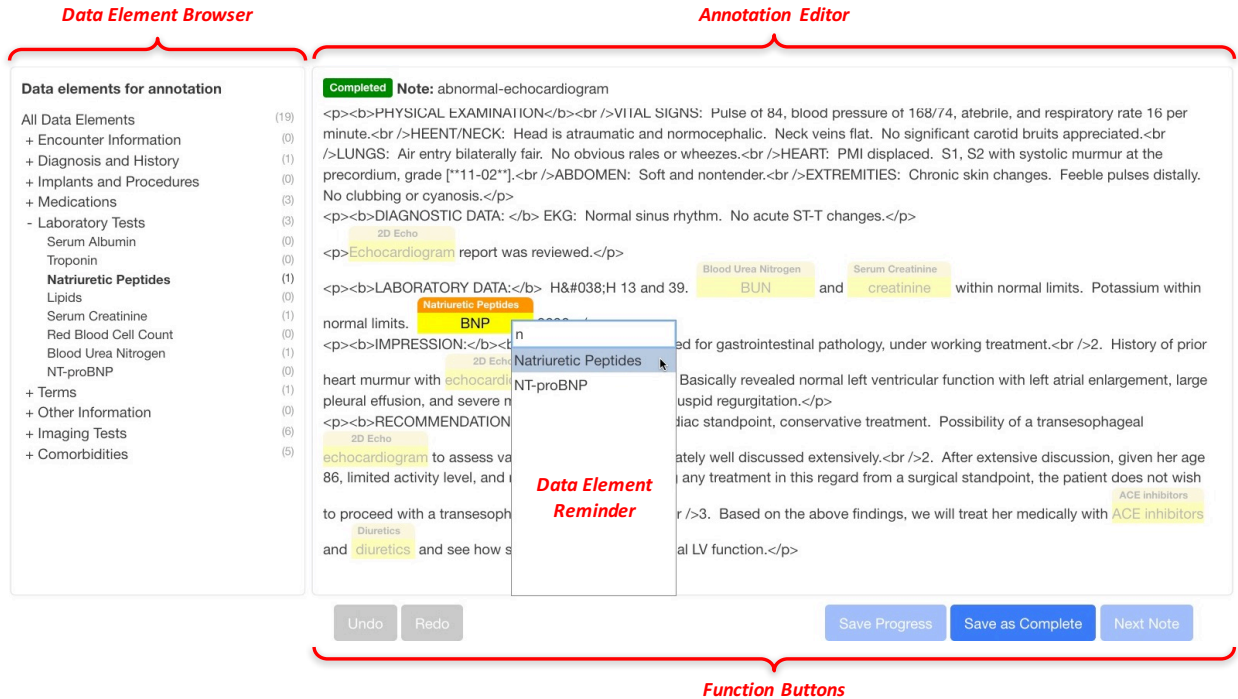

**Figure 2.** The CLEAN annotation tool (CLEAN-AT). The left panel (*data element browser*) shows all the data elements, while the right panel (*annotation editor*) is for annotation adding, deleting, and modifying via the pop-up menu (*data element reminder*). The bottom panel (*function buttons*) allows users to undo/redo editing, save progress, save the note as complete, and proceed with the next note. The clinical note shown here is from publicly available MT Sample notes.[33]

- *Data element reminder (a popup menu while tagging the text)*. After right-clicking on a selected mention in the annotation editor, a popup menu appears with a list of all data elements grouped by the categories. A user can use the keyboard as an auto-completion shortcut to filter possible data elements and thus can speed up the annotation editing process. For the example shown in Figure 2, if the user types "n," CLEAN-AT would display "Natriuretic Peptides" and "NT-proBNP," the data elements of which name starts with "n." The user can then choose from the filtered data elements to annotate or re-annotate quickly.

- *Undo/redo (function buttons in the bottom panel)*. CLEAN-AT supports an unlimited number of undo and redo operations, which allows users to edit and delete quickly, knowing that they could easily recover from mistakes. Note that, deletions are also considered as an operation, and therefore user can undo/redo deletions. Supporting undo/redo operations follows an important principle of user interface design to reduce distress of new users.[34-36] However, few existing annotation tools for cNLP support undo/redo, because the program must record every user interaction. Also, the program must maintain a stack data structure, to save the entire history of previous user interactions that need to be accessible anytime the user attempts to undo/redo the last operations. If the software did not include the undo/redo function in the initial design, it would be difficult to add this feature because the whole program might need to be thoroughly rewritten. CLEAN-AT implements this important feature based on the JavaScript library React.[37]

- Overview of the completion status of clinical notes. As shown in Figure A.2 in Appendix A, the user can start a new annotation process, continue an incomplete annotation process, or recheck a previously completed annotation process, for any clinical note in a target corpus. Although the overview page shows the overall information for all notes, CLEAN-AT brings up the next note for annotation after the user completes the annotation of a clinical note, instead of bringing up this overview page. This feature was designed to effectively keep a user engaged.

### 3.3 Study Material

We evaluated the performance of CLEAN using two cohorts: (1) Congestive Heart Failure (CHF) as an exemplar of a major health condition in the U.S., where the prevalence of heart disease is 5.7 M;[38] and (2) Kawasaki Disease (KD), which represents a rare yet acute disease with an estimated number of U.S. hospitalizations of 5,447 in 2009.[39] Both conditions are use-case conditions for the pSCANNER clinical data research network.[22] We used the clinical note pool and target data elements described in our previous work [1] with newly annotated gold standards, which are further described later in this subsection.

Our pool of clinical notes contained notes from public domain datasets,[1] including MT Samples,[33] i2b2 Challenges 2009 – 2012 and 2014,[40-51] ShARe CLEF eHealth Tasks 2013 Task 1 and 2 (which are now part of the MIMIC III clinical database [52]) and 2014 Task 1.[53-55] We collected the corpus of notes for CHF or KD by selecting notes that contained "*congestive heart failure*" or "*Kawasaki* OR (*fever* AND *rash* AND *red* AND *child*)," respectively. Because KD is a relatively rare disease,[8 39] we included MT Sample clinical notes containing "*fever*" to increase the corpus size for KD. The resulting corpus consisted of 635 notes for CHF and 33 notes for KD.

In our previous study,[1] CHF and KD subject matter experts had identified the target data elements, which were then mapped to standardized concept IDs defined in SNOMED-CT,[56] LOINC,[57] RxNorm,[58] or UMLS.[59] After mapping, the normalized output data elements of the cNLP tools were ready to serve as inputs to the CLEAN-EP ensemble pipeline.

For each clinical note, an experienced physician annotated the mentions of each data element as the gold standard. During the annotation process, the physician followed the cNLP annotation guidelines we developed. The physician used the CLEAN-EP pre-annotations and the CLEAN-AT annotation tool to annotate the gold standard data elements.

**3.4 Study Design**

The target users of CLEAN are clinical researchers, front-line clinicians, and their supporting staff members, who would like to identify a cohort of patients from an EHR system for their scientific and/or quality improvement projects. Therefore, our inclusion criteria for a test user included: (1) an employee or a student of UCSD who was $\geq$ 18 years old, and (2) who had worked or was soon going to work on cohort identification using annotated clinical notes. Based on these criteria, two approved test users with required training certificates (HIPPA[1] and CITI[2]) were selected to participate in our study: a practicing physician with clinical training (a domain expert), and a graduate student in the Department of Biomedical Informatics with basic biomedical knowledge (a domain novice). Our annotation tasks involved keyboard, mouse, audio and video recording to precisely log all interactions, with timestamps, between the test user and the software. The Institutional Review Board (IRB) at UCSD approved this study with Project Number 160410 on April 21, 2016.

Two-sample paired t-test was used to compare the mean difference in correctness and efficiency measured with two systems on the same clinical notes. The assumed effect size was 0.5, computed from mean of the correctness metric (F1-score in our study) (0.8 and 0.7 for CLEAN and BRAT, respectively), and standard deviation 0.2 for both systems, based on the experimental results of our previous work.[1] The estimated statistical power was 78%, with effect size 0.5, significance level 0.05 and per-group size 32.[60] Our study had the required sample size to evaluate CLEAN and BRAT, with significant statistical power, at least 32 clinical notes for each cohort.

Based on the estimation of sample size, Figure 3 illustrates our process to select the clinical notes in the test datasets from the corpus of notes, which contained 635 notes for CHF and 33 notes for KD. First, 32 KD notes were randomly selected, such that the notes could be divided equally between CLEAN and BRAT. Then, 52 clinical notes for the CHF cohort were randomly selected, for a total of 84 clinical notes as the test dataset for our comparative studies (Table 1).

---

[1] Health Insurance Portability and Accountability Act
[2] Collaborative Institutional Training Initiative

The subject matter experts had enumerated 87 data elements for the CHF (50) and KD (37) cohorts.[1] Within the 84 test notes, the experienced physician annotated 1,542 mentions of these 87 data elements as our gold standard. We split the notes in the test dataset into 6,702 sentences using the Stanford CoreNLP library.[61] The statistics of the test dataset are shown in the last row of Table 1.

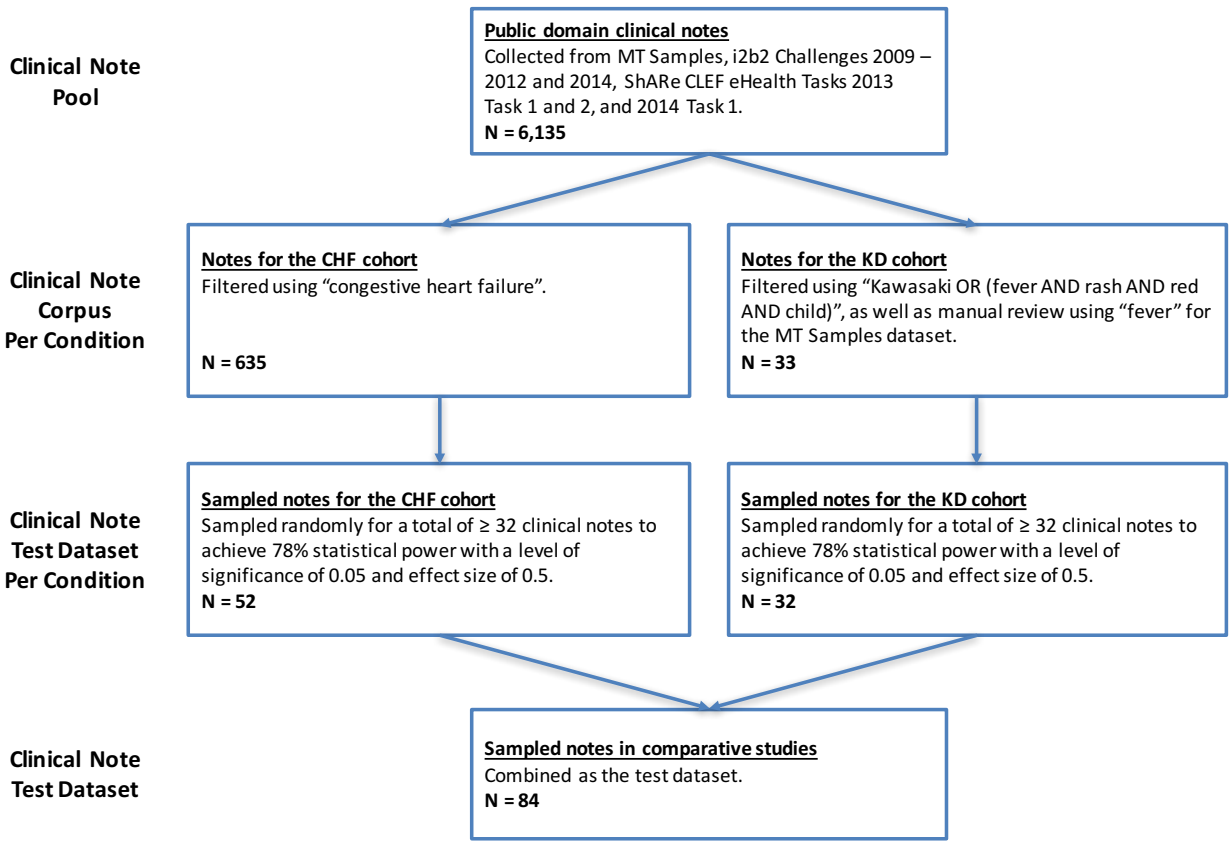

**Figure 3.** The clinical note test dataset selection process for *congestive heart failure (CHF)* and *Kawasaki disease (KD)* conditions.

**Table 1**. Statistics of the total test dataset (the last row) and the two test sets (ID = 1 and 2) of clinical notes. One test set (ID = 1) was used to evaluate BRAT, while the other (ID = 2) was used to evaluate CLEAN. Concept frequency indicates the total number of gold standard annotations in the clinical notes in a test set.

| Test Set ID | Evaluated Software | Notes | Words | Average Words per Note | Standard Deviation Words per Note | Concept Frequency |
|---|---|---|---|---|---|---|
| 1 | BRAT | 42 | 44,219 | 1,053 | 676 | 724 |
| 2 | CLEAN | 42 | 48,998 | 1,167 | 706 | 818 |
| Total (Test Dataset) | | 84 | 93,217 | 1,110 | 689 | 1,542 |

**3.5 Evaluation Protocol**

The 84 clinical notes in the test dataset were randomly divided into two test sets, each with 42 notes (26 CHF and 16 KD). One test set was used to evaluate BRAT, while the other was used to evaluate CLEAN. Table 1 describes the statistics of the two test sets.

Both users evaluated each software with the same test set and same order of clinical notes, but the order of the software evaluated was randomly assigned. The randomization assigned user-1 to annotate the notes using CLEAN first and then BRAT for CHF (but BRAT first for KD), while user-2 annotated the notes using BRAT first then CLEAN for CHF (but CLEAN first for KD).

Each evaluation session took about 1.5 to 2 hours long. Before the annotation, both users attended a 20-minute training session on the cNLP annotation guidelines, data element tables, and the usage of each annotation software. The users could rest at any time during the annotation process or completely cease the session as needed.

**3.6 Evaluation Measurements**

The comparative study of CLEAN and BRAT exploited three types of measurements: correctness, efficiency, and usefulness/satisfaction, which are discussed in Section 3.6.1, 3.6.2, and 3.6.3, respectively.

### 3.6.1 Correctness Measurements

The gold standards allowed us to assess the quality of the user-reviewed annotations. For this purpose, our evaluation metric was F1-score, computed under note-level and sentence-level. The detailed definitions of the note-level and sentence-level F1-scores are described in Appendix A.2.

Two mixed-effects models were considered for each response variable of the note- and sentence-level F1-score, with the computation of intra class correlation (ICC). Stepwise backward elimination method was applied to select variables among a list of candidate variables consisting of software (CLEAN versus BRAT), condition (CHF versus KD), length (word count of the clinical note), and concept frequency (number of the gold standard annotations in the clinical note), while having reviewer (graduate student versus physician) as a random effect, with a significance level of 0.05.

### 3.6.2 Efficiency Measurements

The main metric of efficiency was the *average annotation time*, in minutes, to finish an annotation task for each clinical note. A mixed-effects model, similar to the models described in Section 3.6.1, was considered for each response variable of the annotation time, with the computation of ICC. In addition, we considered the total number of keyboard presses and mouse clicks to better understand the *user activities* required to operate the software. The keyboard/mouse interactions and the audio/video recordings were logged using the RUI tool [62

63] and the OBS Studio tool,[64] respectively. Based on our assumption that the keyboard presses and mouse clicks were proportional to the *length* of a clinical note, counts were normalized per word to serve as our metric for user activities.

3.6.3   Usefulness/Satisfaction Measurements

This measurement consisted of scale surveys and qualitative interviews. The *usability* scale survey was a 7-point (1 – 7 Likert scale, 7 being best) questionnaire,[65] with 6 questions related to *Perceived Usefulness* and 5 questions related to *Perceived Ease of Use*. The *satisfaction* scale survey was a 10-point (0 – 9 Likert scale, 9 being best) questionnaire,[66] with 6 questions for *Terminology and System Information*, 6 questions for *Learning*, 6 questions for *Overall Reaction to the Software*, 4 questions for *Screen*, and 5 questions for *Terminology and System Information*. After each annotation session, the users filled out the surveys on the system, and expressed any comments they had about critical events, such as not being able to find a button, in an interview meeting.

## 4. RESULTS

### 4.1 Correctness Results

The results of the correctness comparison of the two systems are shown in Table 2. In general, CLEAN improved the precision, recall and F1-score for both note- and sentence-level evaluations. The comparison results using boxplots (Figure 4 (a) and (b)) further indicated that CLEAN improved the correctness significantly, with resulting P-values <0.001 and 0.004 for note- and sentence-level, respectively.

The F1-scores of the cNLP pre-annotation are given below. For note-level, the cNLP pre-annotation F1-scores were 0.695 for the clinical notes in BRAT and 0.791 for the notes in CLEAN. For sentence-level, the cNLP pre-annotation F1-scores were 0.529 for the clinical notes in BRAT and 0.500 for the notes in CLEAN. That is, the F1-score of CLEAN improved from 0.791 to 0.896 in note-level, and from 0.500 to 0.719 for sentence-level, after the user review via the CLEAN-AT. Note that, in our experiments, CLEAN exploited the cNLP pre-annotated mentions while BRAT did not. It should also be noted that the KD clinical notes included an outlier data point, which generated zero precision, recall and F1-score and was replaced by using interpolation.

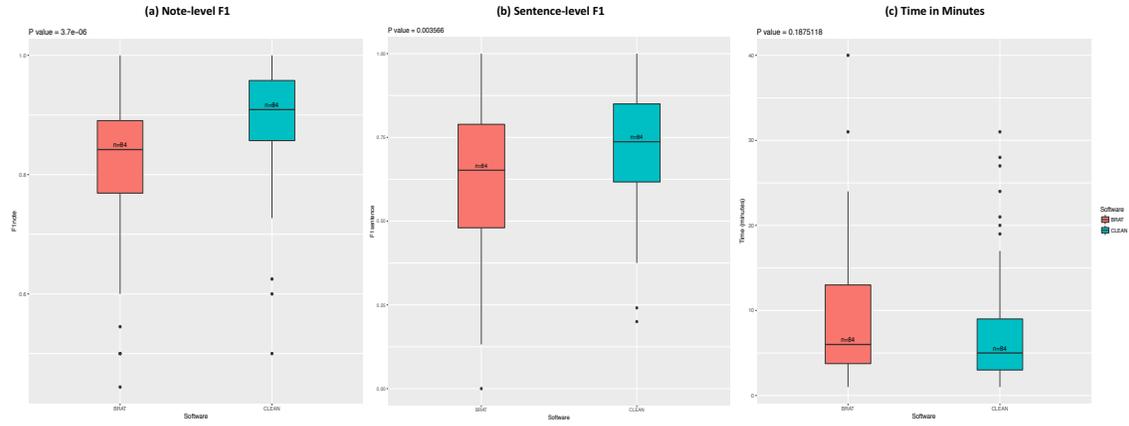

**Figure 4**. (a) Boxplot for the *note-level F1-score*. (b) Boxplot for the *sentence-level F1-score*. (c) Boxplot for *annotation time*.

**Table 2**. Comparison of the gold standard with annotation results from BRAT and CLEAN. The metrics were note-level and sentence-level averaged precision, recall and F1-score, with 95% confidence interval (95% CI).

| Level | Software | Precision (95% CI) | Recall (95% CI) | F1-score (95% CI) |
|---|---|---|---|---|
| Note | BRAT | 0.855 (0.826 to 0.885) | 0.816 (0.780 to 0.852) | 0.820 (0.794 to 0.846) |
| | CLEAN | 0.895 (0.870 to 0.920) | 0.913 (0.890 to 0.936) | 0.896 (0.876 to 0.916) |
| Sentence | BRAT | 0.630 (0.582 to 0.677) | 0.614 (0.565 to 0.664) | 0.616 (0.568 to 0.665) |
| | CLEAN | 0.727 (0.689 to 0.764) | 0.723 (0.684 to 0.763) | 0.719 (0.681 to 0.757) |

Finally, the intra class correlation (ICC) within an annotator was 0.263 and 0.272 for note- and sentence-level F1-score, respectively. The final mixed-effects models showed that both levels of F1-score were mostly related to software differences, with a p-value < 0.001. The note-level F1-score was also related to concept frequency and condition. The details of the mixed-effects modeling results are shown in Table 3. The model plots of note- and sentence- level F1-score are shown in Figure 5 and A.3 in Appendix A, respectively. Under note-level, the F1-score improvement of CLEAN over BRAT was more salient and the p-value was also much lower.

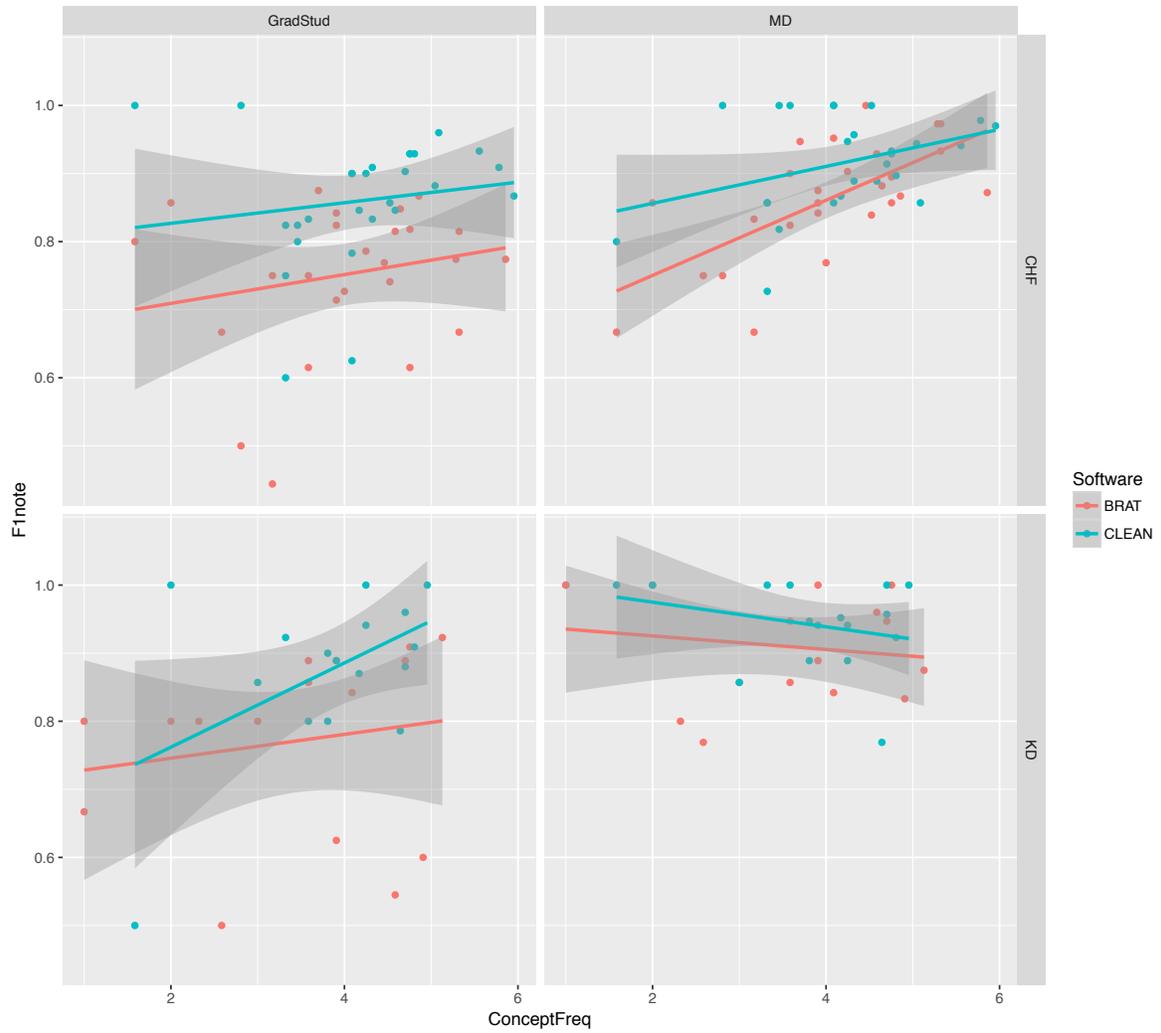

**Figure 5**. The results of the linear mixed-effects model using *note-level F1-score* as response and annotators as random effect. The software, concept frequency and condition effects are included according to the highest relevancy shown in Table 3.

**Table 3.** Final linear mixed-effects model using note- and sentence-level F1 as response, and annotators as random effect. The most relevant factors are shown in bold text.

| Level | Effect | Estimate | Standard Error | P-value |
|---|---|---|---|---|
| Note | (Intercept) | 0.725989 | 0.043538 | 0.000001 |
| | **Software** | **0.070426** | **0.014759** | **0.000004** |
| | Concept Frequency | 0.020996 | 0.007027 | 0.003235 |
| | Condition | 0.037190 | 0.015439 | 0.017095 |
| Sentence | (Intercept) | 0.615911 | 0.061387 | 0.006664 |
| | **Software** | **0.102887** | **0.028570** | **0.000418** |

## 4.2 Efficiency Results

The average annotation time in minutes-per-note for BRAT was 8.286 (with 95% Confidence Interval from 6.788 to 9.783), while the time for CLEAN was 7.262 (with 95% Confidence Interval from 5.859 to 8.665). The comparison results using boxplots (Figure 4 (c)) showed that CLEAN improved the annotation time, but not significantly (p = 0.188).

Also, the ICC within an annotator was 0.418. The final mixed-effects model indicated that annotation time was mostly related to clinical note length with P-value < 0.001. The second most significant effect was the software difference with P-value = 0.013. Furthermore, the annotation time was related to condition and concept frequency. The details of the model are shown in Table 4 and Figure 6, while the detailed time analysis results are shown in Table A.1 in Appendix A.

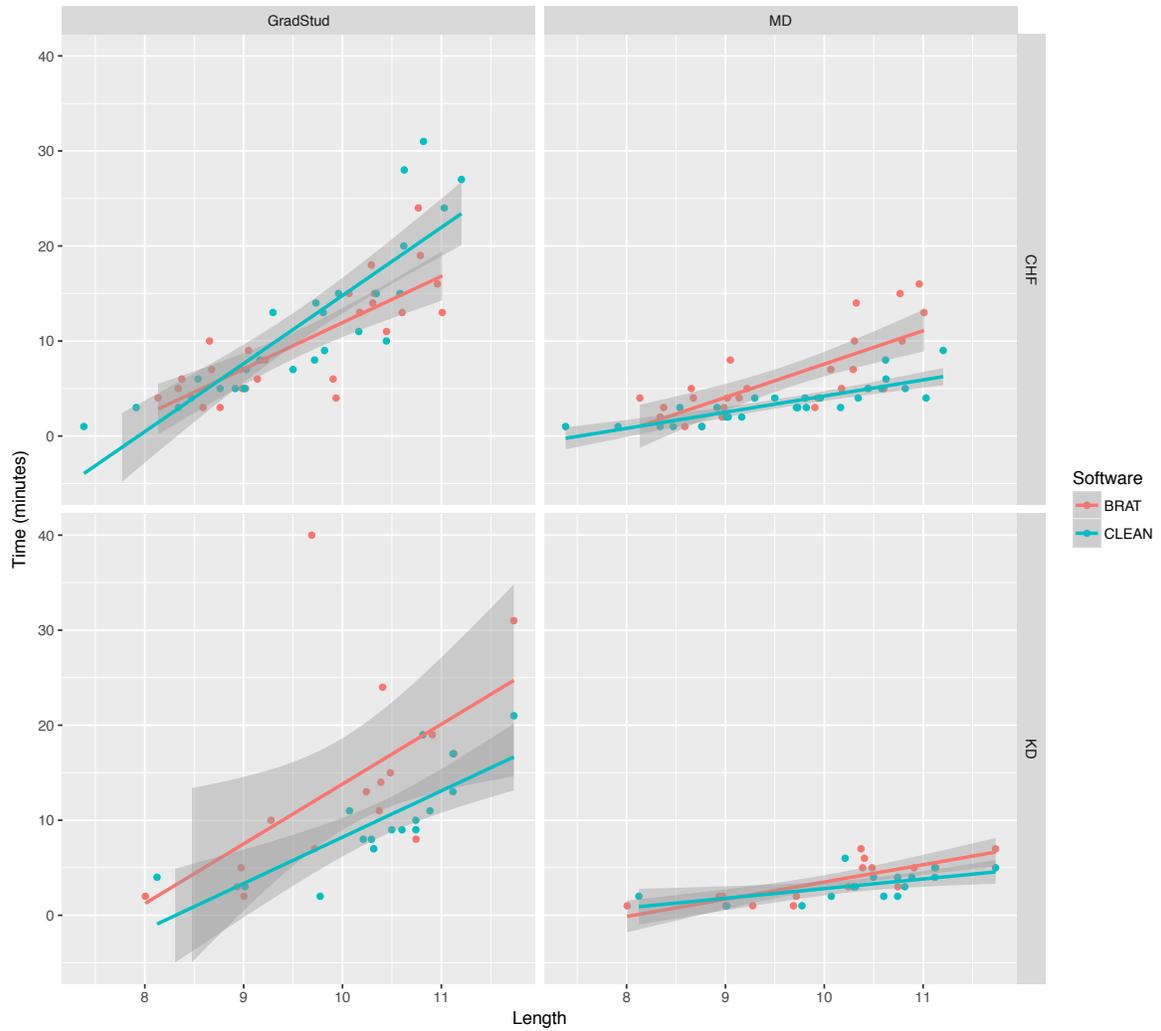

**Figure 6**. The results of the linear mixed-effects model using *annotation time* as response and annotators as random effect. The length, software, and condition effects are included according to the highest relevancy shown in Table 4.

**Table 4.** Final linear mixed-effects model using annotation time as response and annotators as random effect. The top relevant factors are shown in bold text.

| Effect | Estimate | Standard Error | P-value |
|---|---|---|---|
| (Intercept) | -26.888250 | 4.491174 | 0.000007 |
| **Software** | **-1.716550** | **0.685926** | **0.013298** |
| **Length** | **3.346265** | **0.482188** | **0.000000** |
| Concept Frequency | 0.851998 | 0.429080 | 0.048720 |
| Condition | -1.812721 | 0.832729 | 0.030905 |

The statistics of user activities, including keyboard presses and mouse clicks, are shown in Table A.2 in Appendix A. The averaged-and-normalized user activities for BRAT was 0.094 (or 10.6 words reviewed per press/click), and for CLEAN, it was 0.076 (or 13.2 words reviewed per press/click). Therefore, CLEAN reduced both normalized mouse clicks and normalized keyboard presses. Also, Figure A.4 in Appendix A illustrates the results per annotator, showing that CLEAN decreased the normalized user activities compared to BRAT for both annotators.

**4.3 Perceived Usefulness and User Satisfaction Results**

The summary of the survey results is shown in Table A.3 in Appendix A, while the full survey results are shown in Figure A.5 in Appendix A. The interview results, including critical events and overall feedback of BRAT and CLEAN, are described in Appendix A.3. Our findings based on the questionnaire results and interviews are as follows:

- *Perceived Usefulness and Ease of Use*. The physician annotator found superior usefulness of CLEAN over BRAT, with average difference of 3.600 in satisfaction score values over two categories, and the graduate student annotator reported insignificant results with average difference of 0.215.

- *User Interface Satisfaction*. While the physician annotator perceived a higher level of satisfaction of CLEAN compared to BRAT, with average difference of 4.204 in satisfaction score values over five categories, the graduate student annotator showed a modest difference (0.364).

The advantages of BRAT reported by the annotators included low learning curve of the tool, the ability to easily navigate to next notes to annotate, and the auto-save feature of the work progress. However, the BRAT interface showed inefficiency as a system: the feedback speed to user interaction was slow and finding a data element for annotation required many clicks. The participants had to click multiple times to select a text, often giving unwanted text selection, making them cancel the operation and restart the task. The text position often shifted after annotations, making it difficult for the participants to reorient where in the annotation process they were working on.

The reported benefits of CLEAN included quick learning curve and ease of use, similar to BRAT. However, the participants saw pre-annotation, which was a unique feature to CLEAN, to be helpful and facilitating of their annotation process. The tool allowed easy addition or deletion of annotations, providing immediate feedback to each user interaction, leading to efficient annotation tasks. The participants also saw CLEAN to have a potential to be used as a predictive analysis tool. The participants also noted the improvements that can be made with CLEAN as it: allows single-click word selection, has more keyboard-assisted data element selections, enables a font size adjustment feature, and includes more medication synonyms.

## 5. DISCUSSION

Based on the results from our comparative study, we found that, for the two testers, CLEAN could improve correctness (precision, recall and F1-score) of annotation and increase annotation quality. Especially, CLEAN improved note-level F1-score according to the corresponding mixed-effect model. CLEAN can also remained at the same level of annotation efficiency. There was no significant difference in the note-level precision, as the confidence intervals of BRAT and CLEAN were overlapping. However, for the sentence-level, the difference in precision was significant. Also, perceived efficiency of the tool was higher for CLEAN than for BRAT according to the interview results. Therefore, for high-granularity annotation tasks that look into each sentence, CLEAN can help the annotators provide more precise annotation results.

CLEAN could also lower the number of required user activities and decrease the annotation time for a physician annotator, but the time decrement was not significant in general. From the surveys and interviews, the physician annotator reported more usefulness/satisfaction for CLEAN compared to the graduate student. We hypothesize that this may be the case that the student annotator needs to spent substantial amount of time looking upon potential keyword categories and confirming many of cNLP pre-annotation mentions.

One limitation of our study is that we did not test how the pre-annotation correctness rate affects a user's experience. However, we predict that with more accurate cNLP pre-annotation results, the user experience and annotation quality will improve. In addition to that the small number of notes, our experiment included only one user of each type (domain expert or novice) and thus these results may not be generalized well to all users within a type, such as all graduate students. Finally, our current gold standard annotations were created by one annotator and may require further validation.

## 6. CONCLUSION

This study evaluated the CLEAN cNLP system, which included CLEAN-EP to automatically generate pre-annotations and CLEAN-AT for annotators to review the machine-generated pre-annotations. The study compared CLEAN with BRAT and found that CLEAN demonstrated improved correctness and better usefulness/satisfaction, especially for a physician annotator, while retaining the same level of efficiency. CLEAN could help address the bottleneck of annotation in the cNLP concept extraction pipeline to support phenotyping and cohort identification.

**ACKNOWLEDGEMENT**

Part of the de-identified clinical records used in this research were provided by the i2b2 National Center for Biomedical Computing funded by U54LM008748 and were originally prepared for the Shared Tasks for Challenges in NLP for Clinical Data organized by Dr. Ozlem Uzuner, i2b2 and SUNY. The computational infrastructure was provided by the iDASH National Center for Biomedical Computing funded by U54HL108460 and managed by the Clinical Translational Research Institute CTSA Informatics team led by Antonios Koures, PhD, funded in part by UL1TR001442.

**FUNDING STATEMENT**

Tsung-Ting Kuo, Jina Huh, Robert El-Kareh, Gordon Lin, Michele E. Day, Lucila Ohno-Machado, and Chun-Nan Hsu are partially funded through a Patient-Centered Outcomes Research Institute (PCORI) Award (CDRN-1306-04819). The statements in this article are solely the responsibility of the authors and do not necessarily represent the views of PCORI, its Board of Governors or Methodology Committee. Part of the de-identified clinical records used in this research were provided by the i2b2 National Center for Biomedical Computing funded by U54LM008748 and were originally prepared for the Shared Tasks for Challenges in NLP for Clinical Data organized by Dr. Ozlem Uzuner, i2b2 and SUNY. The computational infrastructure was provided by the iDASH National Center for Biomedical Computing funded by U54HL108460.

**COMPETING INTERESTS STATEMENT**

The authors have no competing interests to declare.

**CONTRIBUTORSHIP STATEMENT**

T-TK designed and implemented the system, conducted literature review, collected the data, developed the annotation guideline, provided training sessions, performed experiments, analyzed the results, and drafted the manuscript. JH provided feedbacks on the study and system design, suggested critical directions for efficiency, usability and satisfactory evaluations, performed experiments, and edited the manuscript. JK provided feedback on the study design, conducted sample size estimation, performed the mixed-effects model analysis, provided insights to present the results, and edited the manuscript. RE-K provided feedback on the study and system design, developed the annotation guideline, annotated the gold standards, provided critical discussion points, and edited the manuscript. SS and SFF provided feedback on the study and system design, developed the annotation guideline, evaluated the software systems, made suggestions for system efficiency, usability and satisfaction improvement, and edited the manuscript. VK and GL implemented the system and edited the manuscript. MED provided feedbacks on the idea, and edited the manuscript. LO-M was principal investigator of the project; provided overall supervision of the project and critical editing of the manuscript. C-NH provided the original idea, annotation guideline suggestions, critical discussion points, overall supervision of the study and critical editing of the manuscript.

**APPENDIX A**

**A.1 The CLEAN Ensemble Pipeline (CLEAN-EP)**

The CLEAN clinical natural language processing (cNLP) system consists of an ensemble pipeline (CLEAN-EP) as shown in Figure A.1. CLEAN-EP was described in our previous work.[1] The inputs of the pipeline were data elements and clinical notes determined by the study plan, while the outputs were the pre-annotated clinical notes stored on a virtual machine (VM) ready for review. The pipeline consists of four steps:

- *Preprocessor*. CLEAN-EP converts the clinical notes to plain text, transforms them to UTF-8 encoding, and splits them into sentences by using the Stanford CoreNLP library.[61]

- *Toolkit*. CLEAN-EP included three general-purpose cNLP tools: (1) *cTAKES* (clinical Text Analysis and Knowledge Extraction System),[2] an cNLP tool for information extraction from free text clinical notes in EHR; (2) *MetaMap*,[3] a tool for recognizing UMLS concepts in text; and (3) *MedEx*,[4] a tool specialized in extracting mentions of medications. These tools cover a wide range of applications of cNLP for clinical note information extraction and serve the basic needs of clinical note processing. CLEAN-EP also integrated two specialized cNLP tools for pSCANNER [22] conditions: *KD-NLP*,[8] a tool specialized in identifying clinical signs of KD; and *ROCK* (Rules for Obesity, Congestive heart failure, and Kawasaki disease), a newly developed cNLP tool which is specialized in extracting common data elements for WM/O,

CHF, and KD. The pipeline of ROCK consisted of sentence splitting, rule-based tagging using regular expressions, negation detection with NegEx [67], and synonym identification using SNOMED-CT [56] synonyms, LOINC [57] component names, RxNorm [58] trade names, as well as known names for KD found on four public domain medical websites: WebMD [68], MedScape [69], RxList [70], and Drugs.com [71]. Note that, *cTAKES* exploits the Dictionary Lookup Fast Pipeline [72] and the built-in concept dictionary, which is a subset of UMLS [59] including SNOMED-CT [56], RxNorm [58], and all synonyms. It should also be noted that our previous study only included *cTAKES* and *MetaMap*,[1] while CLEAN-EP currently integrates five cNLP tools: *cTAKES*, *MetaMap*, *MedEx*, *KD-NLP* and *ROCK*.

- *Ensembler*. CLEAN-EP adopted the *Union* ensemble method, reported to consistently outperform a single cNLP tool.[1] The *Union* method focuses on improving the coverage of the cNLP results, by extracting a data element if at least one of the cNLP tools identifies that data element.[1]

- *Postprocessor*. CLEAN-EP generates the pre-annotations in the format of BRAT [21] for the evaluation processes.

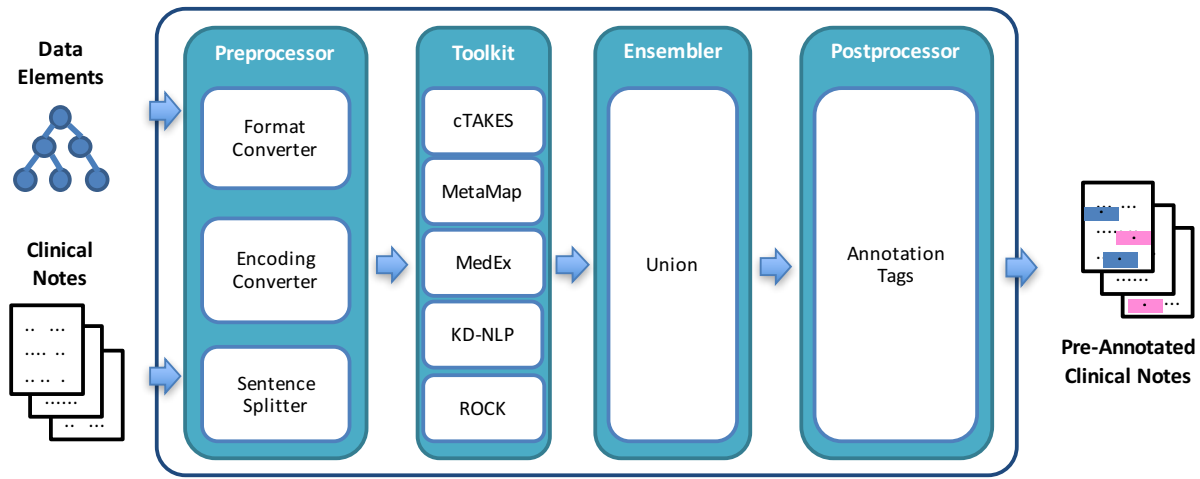

**Figure A.1** The flowchart of the *CLEAN ensemble pipeline (CLEAN-EP)*.[1] The inputs were data elements and clinical notes, while the outputs were the pre-annotated clinical notes. The original ensemble pipeline [1] contained two cNLP tools: *cTAKES* [2] (clinical Text Analysis and Knowledge Extraction System) and *MetaMap*.[3] To increase the extraction coverage, CLEAN-EP integrated three additional cNLP tools: *MedEx*,[4] *KD-NLP*,[8] and a newly developed tool *ROCK* (Rules for Obesity, Congestive heart failure, and Kawasaki disease). Also, CLEAN adopted the *Union* ensemble method.

**Figure A.2.** An example showing the completion status of clinical notes in the *CLEAN annotation tool (CLEAN-AT)*. A user could recheck completed notes (in this example, the upper three notes) by clicking the orange "Recheck" buttons, and could review uncompleted notes (in this example, the lower five notes) by clicking the blue "Review" buttons.

## A.2 The Note-Level and Sentence-Level F1-Scores

At *note-level*, an annotated mention of a data element was regarded as a true positive if the data element appeared in the gold standard annotations for the same *note*. At *sentence-level*, an annotated data element mention was a true positive if it appeared in the same sentence as any gold standard annotation for the data element. In our experiment, data element annotations were considered binary at both levels, and therefore multiple annotations of a data element would only be counted as a true positive. For example, consider the following sentence: "The patient is on *beta-blockers* and Coumadin, continue *beta-blockers*." Although both mentions of the medication concept "*beta-blockers*" are correctly annotated, the true positive count is still one while computing the sentence-level F1-score. The same rule also holds for the computation of note-level F1-score. The definition of *note-level F1-score* of a clinical note was the harmonic mean of precision $P$ and recall $R$ formulated as $2 * P * R / (P + R)$, where $P$ = (# of true positive annotated data elements) / (# of unique data elements annotated in the note), and $R$ = (# of true positive annotated data elements) / (# of unique gold standard data elements in the note). The definition of *sentence-level F1-score* of a clinical note was $\Sigma(F_i) / N$, where $N$ = # of sentences in the clinical note, $F_i = 2 * P_i * R_i / (P_i + R_i)$ for each sentence $i$, $P_i$ = (# of true positive annotated data elements in sentence $i$) / (# of unique data elements annotated in sentence $i$), and $R_i$ = (# of true positive annotated data elements in sentence $i$) / (# of unique gold standard data elements in sentence $i$).

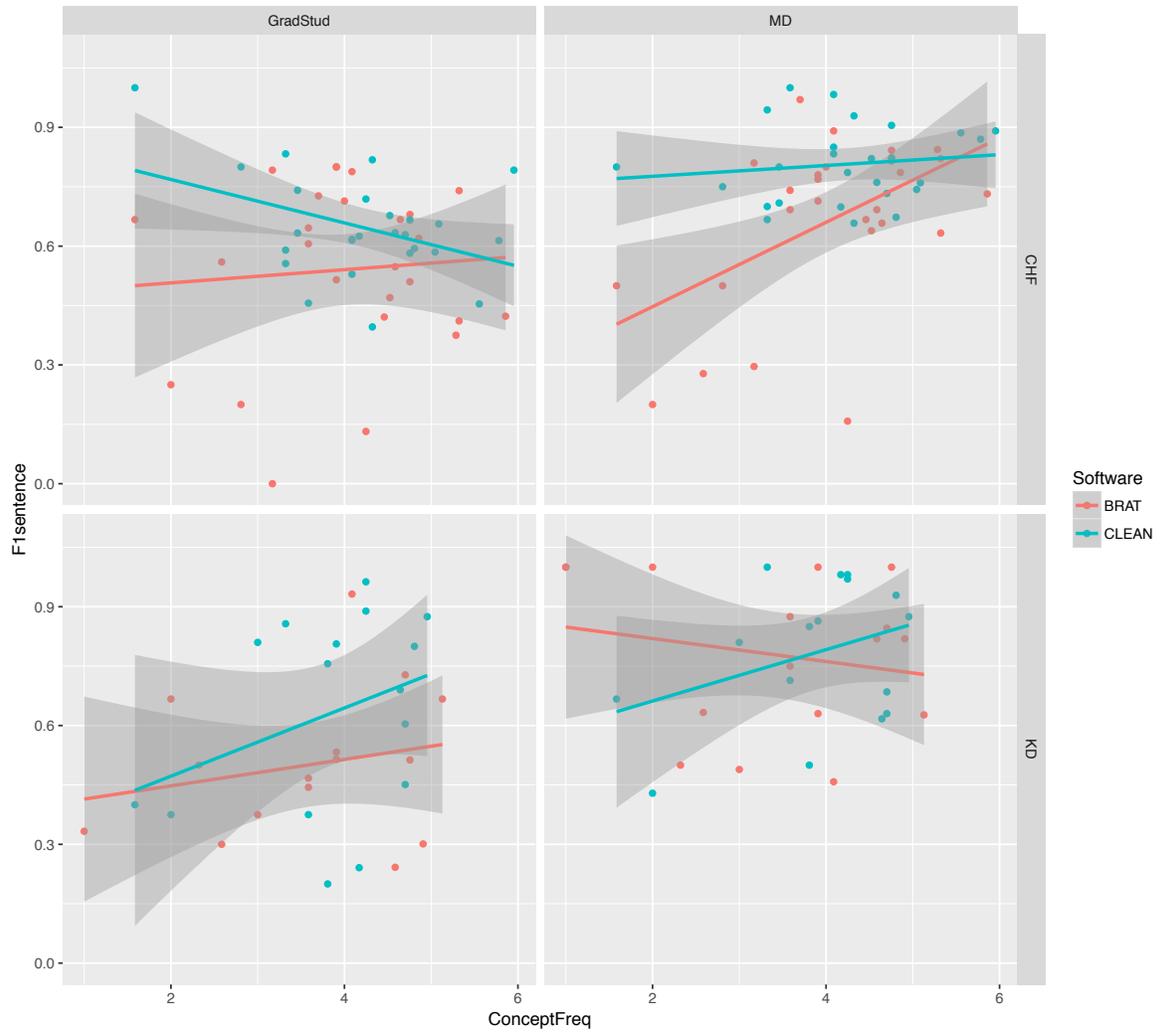

**Figure A.3**. The results of the linear mixed-effects model using *sentence-level F1-score* as response and annotators as random effect.

**A.3 The Interview Results**

The critical events and overall feedback of BRAT and CLEAN are as follows:

- *Interview for critical events of BRAT*. The reported/observed issues are: (1) no pre-annotations as "red flags" existed to assist the annotation process; (2) users tended to highlight a term (or even a large paragraph of the text) accidentally, and thus required an additional cancel operation; and (3) if users zoomed the browser window in for larger text, then the browser cut the pop up window at the bottom.

- *Interview for critical events of CLEAN*. The reported/observed issues are: (1) it constantly pre-annotated some terms as the wrong data element, for example all "history" mentions were pre-annotated as "past medical history," thus the users were required to repeatedly delete/fix these data elements; (2) the coverage of medication pre-annotation could be improved; and (3) based on current design, the copy-and-paste did not work well while trying to query the concepts via search engines like Google.

- *Interview for overall feedback of BRAT*. The annotators reported the following advantages of BRAT: (1) it was easy to learn; (2) it was simple and straightforward to understand, without too many bells and whistles; (3) it provided navigation to next notes with simple clicking; and (4) it provided auto-saving feature. The annotators reported the following burdens of BRAT: (1) the processing was slow; (2) the users needed to click many times to select annotation text; (3) after annotation, the screen did not remain at the site of annotation, but rather shifted back; and (4) searching for data elements, especially medication ones, slowed down the annotation process.

- *Interview for overall feedback of CLEAN*. The annotators reported the following benefits of CLEAN: (1) it was very quick and easy to learn; (2) it did a good job of pre-annotation; (3) adding and deleting an annotation were easy; (4) the user felt more interaction while annotating; (5) it was easy to use, assuming the accurate learning aspect of the software; (6) it was a great tool for annotating clinical notes, especially on a condition basis; and (7) it provided great potential for predictive analysis. The users reported the following obstacles: (1) it would be helpful if with a single-click on a word (instead of clicking-and-selecting the word followed by a right click), that word got highlighted with the annotation data element reminder opened automatically; (2) in selecting the annotation, it might be helpful to allow keyboard use to select an option, once the spelling has started; (3) users could not adjust font size easily; and (4) it might be helpful to add more medication definitions and synonyms to the dictionaries of the cNLP tools to increase the pre-annotation coverage.

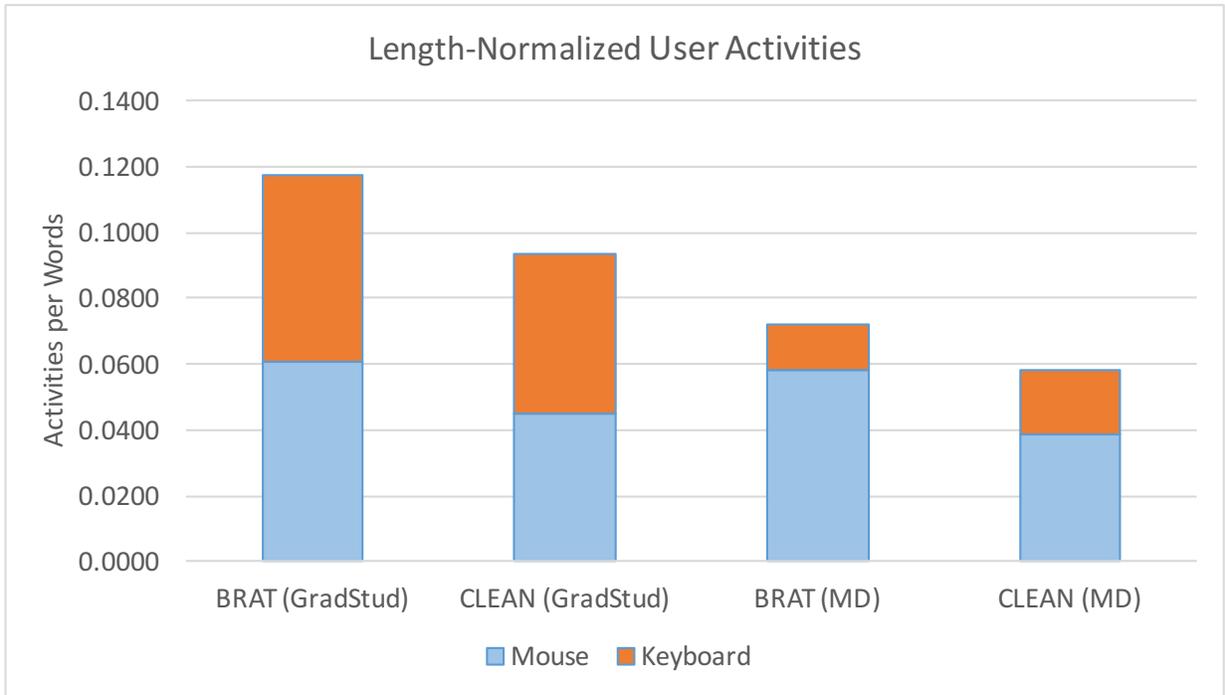

**Figure A.4**. The results of *normalized user activities*, including keyboard presses and mouse clicks, for physician (MD) and graduate student (GradStud) annotators. The activity count was normalized by using the length of clinical note (per word).

| Survey (Scale) | Category | Question | BRAT GradStud | CLEAN GradStud | BRAT MD | CLEAN MD |
|---|---|---|---|---|---|---|
| Usefulness (1 – 7) | Perceived Usefulness | Using the system in my job would enable me to accomplish tasks more quickly | 6 | 7 | 2 | 7 |
| | | Using the system would improve my job performance | 7 | 7 | 2 | 7 |
| | | Using the system in my job would increase my productivity | 7 | 7 | 2 | 7 |
| | | Using the system would enhance my effectiveness on the job | 5 | 7 | 2 | 7 |
| | | Using the system would make it easier to do my job | 6 | 7 | 2 | 7 |
| | | I would find the system useful in my job | 6 | 7 | 2 | 7 |
| | Perceived Ease of Use | Learning to operate the system would be easy for me | 7 | 7 | 5 | 7 |
| | | I would find it easy to get the system to do what I want it to do | 7 | 6 | 4 | 6 |
| | | My interaction with the system would be clear and understandable | 7 | 6 | 4 | 6 |
| | | I would find the system to be flexible to interact with | 7 | 7 | 3 | 6 |
| | | I would find the system easy to use | 7 | 7 | 4 | 6 |
| Satisfaction (0 – 9) | Terminology and System Information | Use of terms throughout system: inconsistent (0) - consistent (9) | 9 | 9 | 4 | 8 |
| | | Terminology related to task: never (0) - always (9) | 7 | 5 | 3 | 8 |
| | | Position of messages on screen: inconsistent (0) - consistent (9) | 9 | 9 | 4 | 8 |
| | | Prompts for input: confusing (0) - clear (9) | 9 | 9 | 3 | 8 |
| | | Computer informs about its progress: never (0) - always (9) | N/A | 0 | 1 | 7 |
| | | Error messages: unhelpful (0) - helpful (9) | N/A | 9 | 2 | 8 |
| | Learning | Learning to operate the system: difficult (0) - easy (9) | 9 | 9 | 5 | 8 |
| | | Exploring new features by trial and error: difficult (0) - easy (9) | 9 | 9 | 3 | 8 |
| | | Remembering names and use of commands: difficult (0) - easy (9) | 5 | N/A | 4 | 8 |
| | | Performing tasks is straightforward: never (0) - always (9) | 9 | 9 | 4 | 8 |
| | | Help messages on the screen: unhelpful (0) - helpful (9) | N/A | 9 | 3 | 8 |
| | | Supplemental reference materials: confusing (0) - clear (9) | 8 | 9 | 3 | 8 |
| | Overall Reaction to the Software | terrible (0) - wonderful (9) | 8 | 8 | 3 | 8 |
| | | difficult (0) - easy (9) | 8 | 9 | 7 | 8 |
| | | frustrating (0) - satisfying (9) | 9 | 9 | 4 | 8 |
| | | inadequate power (0) - adequate power (9) | 8 | 7 | 4 | 8 |
| | | dull (0) - stimulating (9) | 8 | 9 | 3 | 8 |
| | | rigid (0) - flexible (9) | 9 | 9 | 2 | 8 |
| | Screen | Reading characters on the screen: hard (0) - easy (9) | 9 | 9 | 3 | 8 |
| | | Highlighting simplifies task: not at all (0) - very much (9) | 5 | 9 | 3 | 8 |
| | | Organization of information: confusing (0) - very clear (9) | 9 | 7 | 3 | 8 |
| | | Sequence of screens: confusing (0) - very clear (9) | 9 | 9 | 5 | 8 |
| | System Capabilities | System speed: too slow (0) - fast enough (9) | 7 | 9 | 2 | 7 |
| | | System reliability: unreliable (0) - reliable (9) | 8 | 9 | 3 | 7 |
| | | System tends to be: noisy (0) - quite (9) | 5 | 9 | 7 | 9 |
| | | Correcting your mistakes: difficult (0) - easy (9) | 9 | 9 | 3 | 6 |
| | | Designed for all levels of users: never (0) - always (9) | 9 | 9 | 3 | 7 |

**Figure A.5**. Full results of the *usefulness/satisfaction surveys*, completed by the graduate student (GradStud) and M.D. (MD) annotators.

**Table A.1.** Time analysis results (GradStud = Graduate Student, MD = Medical Doctor).

| Condition | Annotator | Software | Average Time per Note (minutes) | Standard Deviation of Time per Note (minutes) |
|---|---|---|---|---|
| Congestive Heart Failure (CHF) | GradStud | BRAT | 9.962 | 5.611 |
| | | CLEAN | 11.808 | 8.266 |
| | MD | BRAT | 6.154 | 4.315 |
| | | CLEAN | 3.500 | 2.045 |
| Kawasaki Disease (KD) | GradStud | BRAT | 13.813 | 10.647 |
| | | CLEAN | 10.063 | 5.360 |
| | MD | BRAT | 3.500 | 2.191 |
| | | CLEAN | 3.188 | 1.471 |

**Table A.2**. User activities for BRAT and CLEAN, averaged for both graduate student and physician users and normalized by using the length of clinical note (per word).

| Software | Mouse Count | Keyboard Count | Total Count | Word Count | Mouse Normalized Count | Keyboard Normalized Count | Total Normalized Count |
|---|---|---|---|---|---|---|---|
| **BRAT** | 2,629.5 | 1,553.0 | 4,182.5 | 44,219 | 0.059 | 0.035 | 0.094 |
| **CLEAN** | 2,057.0 | 1,667.5 | 3,724.5 | 48,998 | 0.042 | 0.034 | 0.076 |

**Table A.3**. Summary of the usefulness/satisfaction survey results for physician (MD) and graduate student (GradStud) annotators. The upper-part summarizes the survey results for perceived usefulness and ease of use, while the lower-part the results for user interface satisfaction. Our study excluded four satisfactory survey questions with N/A answers when computing the average scores for each category of questions.

| Survey (Scale) | Category | BRAT (GradStud) | CLEAN (GradStud) | BRAT (MD) | CLEAN (MD) |
|---|---|---|---|---|---|
| **Usefulness (1 – 7)** | Perceived Usefulness | 6.17 | 7.00 | 2.00 | 7.00 |
| | Perceived Ease of Use | 7.00 | 6.60 | 4.00 | 6.20 |
| **Satisfaction (0 – 9)** | Overall Reaction to the Software | 8.33 | 8.50 | 3.83 | 8.00 |
| | Screen | 8.00 | 8.50 | 3.50 | 8.00 |
| | Terminology and System Information | 8.50 | 8.00 | 3.50 | 8.00 |
| | Learning | 8.75 | 9.00 | 3.75 | 8.00 |
| | System Capabilities | 7.60 | 9.00 | 3.60 | 7.20 |